\pdfoutput=1

\documentclass[11pt]{article}

\usepackage[final]{emnlp2021}
\usepackage{times}
\usepackage{latexsym}
\usepackage{booktabs}

\usepackage[T1]{fontenc}

\usepackage[utf8]{inputenc}

\usepackage{microtype}\usepackage{microtype}

\usepackage{graphicx}

\newcommand{\BB}[0]{BERT\textsubscript{BASE}}
\newcommand{\BL}[0]{BERT\textsubscript{LARGE}}
\newcommand{\RB}[0]{RoBERTa\textsubscript{BASE}}
\newcommand{\RL}[0]{RoBERTa\textsubscript{LARGE}}
\newcommand{\GS}[0]{GPT2\textsubscript{SMALL}}
\newcommand{\GM}[0]{GPT2\textsubscript{MEDIUM}}
\newcommand{\GL}[0]{GPT2\textsubscript{LARGE}}
\newcommand{\GX}[0]{GPT2\textsubscript{XL}}

\title{Sorting through the noise: Testing robustness of information processing in pre-trained language models}

\author{Lalchand Pandia \\
  \texttt{lcpandia@gmail.com} \\\And
  Allyson Ettinger \\
  Department of Linguistics  \\University of Chicago \\
  \texttt{aettinger@uchicago.edu} \\}

\begin{document}
\maketitle
\begin{abstract}
 Pre-trained LMs have shown impressive performance on downstream NLP tasks, but we have yet to establish a clear understanding of their sophistication when it comes to processing, retaining, and applying information presented in their input. In this paper we tackle a component of this question by examining robustness of models' ability to deploy relevant context information in the face of distracting content. We present models with cloze tasks requiring use of critical context information, and introduce distracting content to test how robustly the models retain and use that critical information for prediction. We also systematically manipulate the nature of these distractors, to shed light on dynamics of models' use of contextual cues. We find that although models appear in simple contexts to make predictions based on understanding and applying relevant facts from prior context, the presence of distracting but irrelevant content has clear impact in confusing model predictions. In particular, models appear particularly susceptible to factors of semantic similarity and word position. The findings are consistent with the conclusion that LM predictions are driven in large part by superficial contextual cues, rather than by robust representations of context meaning.  
\end{abstract}

\section{Introduction}

In recent years, pre-trained language models (LMs) have taken NLP by storm. As we work to interpret the impressive performance of these models, a persistent question is the extent to which LMs are doing something like language ``understanding''. Do these models robustly extract the nuances of information conveyed in text, or is their strong performance driven by more superficial mechanisms?

In this paper we focus on examining a particular aspect of ``understanding'' in pre-trained LMs: how robustly these models process, retain, and apply new facts presented in their inputs.  We assume that a fundamental aspect of language ``understanding'' will be the capacity to represent and accumulate information from the meaning of the input text. So if we input ``Sebastian lives in France, and Rowan lives in Indonesia.'', we would expect a model that understands language to form (or update) representations for these imaginary entities Sebastian and Rowan, such that those representations contain the information that Sebastian lives in France, and that Rowan lives in Indonesia. While recent work has studied models as knowledge bases, testing their ability to reproduce facts encountered during training, in this paper we are asking an importantly different question---not about what models memorize during training, but about how sophisticated they are in processing and representing information from new input text after training. 

To test LMs' ability to process and retain information from context, we design cloze tasks that incorporate a piece of critical information in context, and then prompt the model to complete a statement related to that information. To test the robustness of the processes informing these predictions, we introduce distracting but irrelevant content in the contexts, and test whether the models maintain correct predictions in the face of these distractions. Additionally, to explore further the nature of mechanisms informing model predictions, we systematically vary the nature of the distracting content---manipulating how many distractor words we use, how semantically related they are to critical words, and their relative positions in the sentence. 

We apply these tests to a range of recent pre-trained LMs and examine the impacts of our manipulations on model performance. The results indicate clearly that distracting content in the context is effective in undermining model predictions, and variation of distractor types suggests that models are particularly sensitive to influences of semantic similarity and relative word position. Overall, the results support the conclusion that predictions in pre-trained LMs are driven in large part by superficial contextual cues, rather than by robust representations of relevant facts from context. We make all data and code available for further testing.\footnote{\url{https://github.com/lalchand-pandia/Sorting-Through-The-Noise}}

\section{Related Work} 

Prior work has tested LMs as knowledge bases using cloze-style probes~\cite{petroni-etal-2019-language,jiang-etal-2020-know}. As a starting point we rely on models' ability to display this type of knowledge, but our question differs importantly from that work: we are not asking whether models can recall facts about the real world from training---rather, we are trying to gauge the extent to which models form robust representations of new information presented in input after training. Somewhat more similar to ours is work like~\citet{DBLP:journals/corr/abs-2102-01017}, which explores the consistency of models' generation of facts in the face of rephrasing of prompts. The basic intuition behind this work---that LMs' ability to make intelligent-looking predictions can be sensitive to the particulars of the context---is one that we also use as we ask more specific questions about models' processing of information in their input. 

A good deal of prior work has focused on testing for linguistic knowledge in language models ~\cite{rogers-etal-2020-primer}. Much of this work has prioritized testing syntax in pre-trained LMs via agreement tests~\cite{linzen-etal-2016-assessing,gulordava-etal-2018-colorless}. Others expand to broader sets of syntactic phenomena~\cite{wilcox-etal-2018-rnn,futrell-etal-2019-neural, 10.1162/tacl_a_00321} and semantic/pragmatic phenomena~\cite{ettinger-2020-bert}. Other work has studied syntactic and semantic information in contextualized embeddings from these models~\cite{hewitt-manning-2019-structural,tenney2018you,klafka-ettinger-2020-spying}. We take one step up from examination of these abstract linguistic capacities, with a focused examination of models' ability to use such linguistic scaffolding to process and retain new information described in text.

Our use of attractors to test model robustness takes inspiration from use of attractors within syntactic testing contexts~\cite{linzen-etal-2016-assessing,gulordava-etal-2018-colorless}, but we focus on semantic relationships in defining attractors, and use the attractors to investigate different aspects of models' processing. Some scattered work has explored more semantic types of attractors for testing LMs---in particular, there is work looking at whether presence of certain context words will \emph{prime} corresponding targets in context. Such work has experimented with contextual factors like distance between prime and target~\cite{kassner-schutze-2020-negated}, as well as contextual constraint~\cite{misra-etal-2020-exploring}. We build on this existing work with a more systematic exploration of impacts of different types of attractors, and with a more targeted goal of testing models' robustness in processing new facts from context.

In focusing on models' ability to extract, retain, and deploy information conveyed in text, our work also relates to tasks in reading comprehension question answering~\cite{rajpurkar2018know,kovcisky2018narrativeqa, mostafazadeh2017lsdsem, yang2018hotpotqa,richardson2013mctest}. Some such work, like the bAbI dataset~\cite{Weston2016TowardsAQ} and CBT~\cite{Hill2016TheGP}, use insertion of additional material to make the tasks generally more difficult---a tactic that also parallels the related method of adversarial testing ~\cite{jia-liang-2017-adversarial, mccoy2019right,nie-etal-2020-adversarial}. There are important similarities in the questions and strategies of these prior works and ours, but we differ in focusing specifically on information processing in LMs, rather than performance of models supervised for a particular downstream task. Unlike those works, our goal is to shed light on robustness of language ``understanding'', and nature of prediction mechanisms, that arise as a result of LM-based pre-training.

\begin{table*}[ht]
\centering
\begin{tabular}{p{.2\textwidth}|p{.75\textwidth}}
\toprule
\multicolumn{2}{c}{\textbf{Base context}} \\
\midrule
Zero attractor & \textbf{Sebastian} lives in \textbf{France}. The capital of Sebastian's country is \_\_\_\_ \\
  \midrule
\multicolumn{2}{c}{\textbf{Multiple-entity attractor setting}} \\
  \midrule
  B-type attractors & \textbf{Sebastian} lives in \textbf{France}, Rowan lives in Indonesia, and Daniel lives in Chile. The capital of Sebastian's country is \_\_\_\_  \\
  T-type attractors  & \textbf{Sebastian} lives in \textbf{France}, Rowan lives in Jakarta, and Daniel lives in Santiago. The capital of Sebastian's country is \_\_\_\_  \\
  Unrelated attractors & \textbf{Sebastian} lives in \textbf{France}, Rowan drives a car, and Daniel writes poetry. The capital of Sebastian's country is \_\_\_\_ \\
  \midrule
  \multicolumn{2}{c}{\textbf{Single-entity attractor setting}} \\
  \midrule
   B-type attractors & \textbf{Sebastian} lives in \textbf{France}, and has visited Indonesia and Chile. The capital of Sebastian's country is \_\_\_\_  \\
   T-type attractors & \textbf{Sebastian} lives in \textbf{France}, and has visited Jakarta and Santiago. The capital of Sebastian's country is \_\_\_\_ \\
    Unrelated attractors & \textbf{Sebastian} lives in \textbf{France}, drives a car, and writes poetry. The capital of Sebastian's country is \_\_\_\_ \\
 \bottomrule
\end{tabular}
\caption{Example items from dataset. For both multiple-entity and single-entity attractor settings, we give examples in the two-attractor condition---but note that the full dataset varies number of attractors from zero to three. B-type attractors refer to attractors in the same semantic class as the \emph{critical background} word, and T-type attractors refer to attractors in the same semantic class as the \emph{target} word.}\label{tab:data}
\end{table*}

\section{Methods} 

We design our tests in the form of cloze tasks, so as to test the pre-trained LMs in their most natural setting, without interference from fine-tuning. We start from a simple base context, in which the model is given a background fact about an imaginary entity, and then is asked to complete a related statement about the entity. For instance: \\\\
\emph{Sebastian lives in France. The capital of Sebastian's country is $[$MASK$]$}  \\

We will refer to ``France'' here as the \emph{critical background} word, and the correct completion ``Paris'' as the \emph{target} word.  For all of our test items, we establish a baseline competence in our tested models, such that all models successfully prefer the correct target completion over a set of closely related completions (to be outlined shortly) within this simple base context. In this way, we establish that the models have the relevant ``world knowledge'' for this prediction---and then we set aside the issue of world knowledge, to focus on examining robustness of information processing.

Of course, if a model is able to predict ``Paris'' in this base example, this could be attributable to a number of causes. On one hand, it could be taken as evidence that the model was able to store a representation of Sebastian as a resident of France, and then when queried about a related statement, the model was able to make use of that stored information to generate a correct prediction. Alternatively, there may be more superficial explanations for the model's success in this completion: for instance, the model may simply be reacting to the fact that ``France'' was recently mentioned, and now the prompt is asking for a capital. What if ``Indonesia'' had also been mentioned? Would the model still recognize that ``Paris'' is the correct completion? 

To tease apart these classes of explanation, we introduce distracting content in the sentences, and test how this content impacts models' outputs. Following the number agreement literature~\cite{linzen-etal-2016-assessing}, we refer to these inserted items as \emph{attractors}. In a system that robustly represents and retains the critical background information from context, attractor content should not prevent the model from continuing to prefer the correct target completion. If the attractor content does change the models' preferences, then we can infer that more superficial predictive mechanisms are likely at play.

\subsection{Attractor manipulations}

Beyond simply testing whether the model \emph{can} be distracted from giving a correct prediction, we also vary the nature of the attractors so as to better understand the specific mechanisms underlying model predictions. We start by selecting attractors with a semantic relation either to the critical background word (e.g., another country), or to the target word (e.g., another capital). We refer to these as \emph{B-type} and \emph{T-type} attractors, respectively. These semantically related attractors allow us to test the hypothesis that models rely on coarse-grained semantic similarities to inform predictions. If this is the case, then we expect the presence of irrelevant but semantically related material to be particularly disruptive to models' predictions. To contrast with the semantically related attractors, we also include \emph{unrelated} attractors that are not semantically related to the critical background fact or the target.

We present each of these attractor types in two forms. In the first, attractors are listed as additional properties of the key entity (that is, the entity involved in the critical background fact). This allows us to test whether models can sort through different facts about an entity and retrieve the relevant fact for prediction. This is the \emph{single-entity} setting. In the second form, attractors are each associated with a different entity, allowing us to test whether models can form and sort through different entity/property links, to retrieve the relevant fact for prediction. This is the \emph{multiple-entity} setting. 

Table~\ref{tab:data} shows examples of the attractor types and conditions. While the table shows only examples with two attractors, we also vary the number of attractors so as to examine the impact of including more versus less distracting content in context. We vary the number of attractors from zero to three.

\subsection{Dataset construction}\label{sec:datcon}

Because our manipulations require control of numerous variables, we generate our test items synthetically. Synthetic data has limitations, of course, but it also has the important advantage of allowing full control over the nature of the items, so we consider it an important complementary approach to use of naturally-occurring datasets. All sentences generated for our dataset are in English.

The strongest constraint on our design of these items was the need for sets of strongly linked, paired components (e.g., countries and capitals). We need paired items for generating our prediction tasks---for instance, in Table~\ref{tab:data}, we rely on the relationship between countries and capitals---and the relationship in each pair must be strong enough that all models make successful predictions in the base context. Furthermore, we need \emph{sets} of such pairs so as to insert semantically related attractors---of the same type as the critical background word or the target word---in the contexts. 

We identify four item sets that meet our criteria: countries and capitals, professions and associated objects, monuments and associated countries, and sports and associated scoring metrics. We then create templates to support predictions for each of the sets. We test various phrasings for our base contexts, and select those that show optimal performance across models~\cite[c.f.][]{jiang-etal-2020-know}.\footnote{For example, \textit{Sebastian works as a florist . For his job, Sebastian sells [MASK]} is a better query than \textit{Sebastian is a florist . For his job, Sebastian sells [MASK]}).} In keeping with prior LM analysis literature, we define successful prediction in relative terms: models are considered successful on an item if in the base context they assign higher probability to the correct target completion over any of the other target words in the same set (e.g., when ``Paris'' is the correct target, models are considered successful if they prefer ``Paris'' over any other capitals in the set). Appendix Table~\ref{tab:basedata} lists all of the items from our sets, along with their selected base contexts. 

To construct the remainder of the dataset, we start with the base contexts and then sample from attractors of the appropriate types and sets. For additional variety, we select randomly from a sample of six entity names, and we also insert variable amounts of additional semantically unrelated material (\textit{sang in a choir, has a sister}, etc) between the key entity and critical background fact. In all, the dataset includes 40,928 items. In semantically related attractor conditions, multi- and single-entity settings each have 12,896 instances, and in semantically unrelated attractor conditions, multi- and single-entity settings each have 7,568 instances.

\begin{figure*}[ht]
    \includegraphics[width=.9\textwidth]{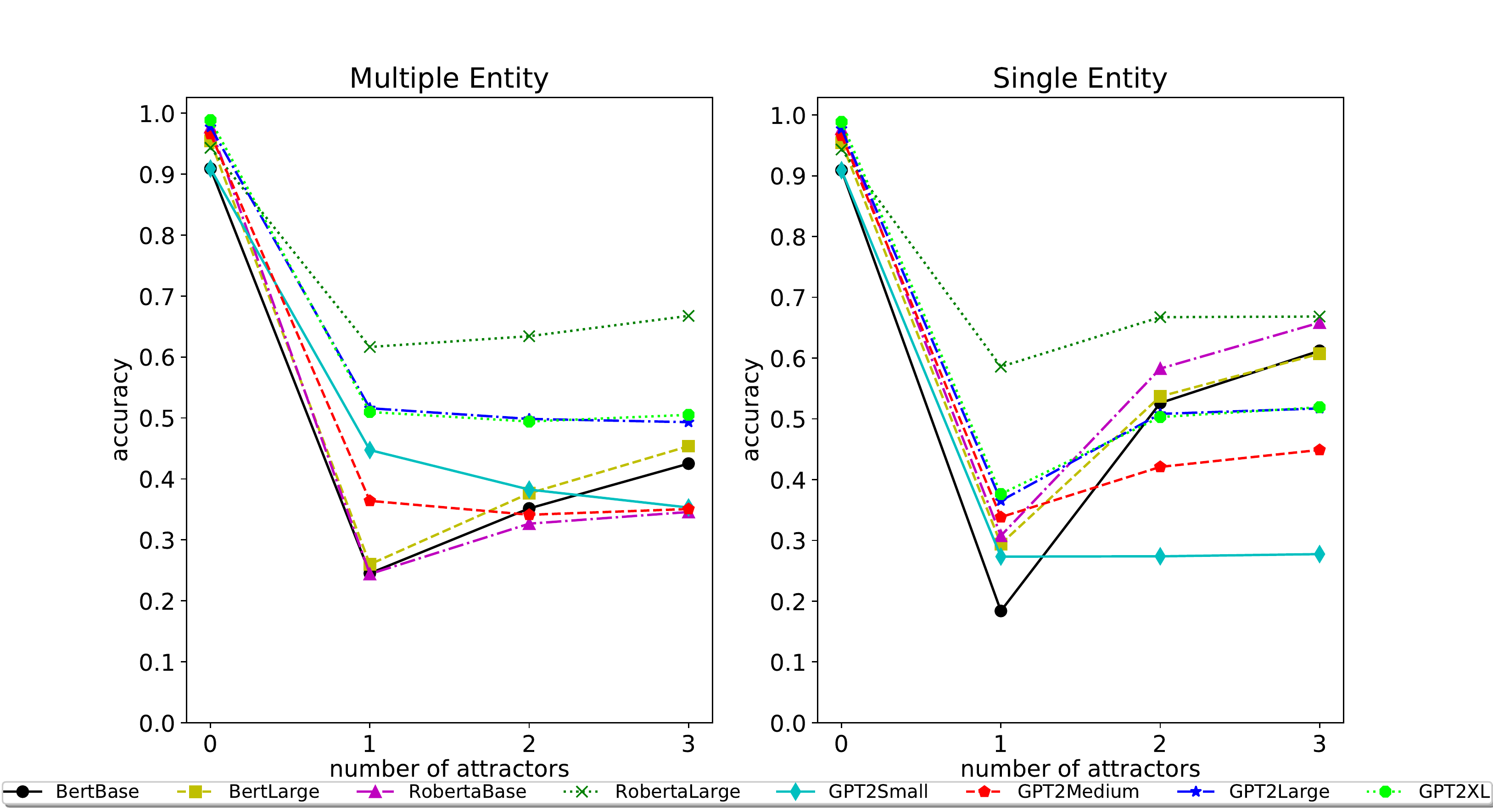}
    \centering
    \caption{Accuracy by number of attractors, with semantically related attractor type}
    \label{fig:accbynum1}
\end{figure*}

\section{Experiments}

\subsection{Models}
We apply our tests to examine three classes of pre-trained LMs, testing various size settings within each class. For the models analyzed in this paper, we use the implementation of~\citet{wolf-etal-2020-transformers}.

\paragraph{BERT~\cite{devlin-etal-2019-bert}} We experiment with two variants: \BB{} (110M parameters), and \BL{} (340M parameters). For both, we use the uncased version.

\paragraph{RoBERTa~\cite{DBLP:journals/corr/abs-1907-11692}} We experiment with \RB{} (125M parameters) and \RL{} (355M parameters).

\paragraph{GPT-2~\cite{radford2019language}} We test \GS{} (117M parameters), \GM{} (345M parameters), \GL{} (774M parameters) and \GX{} (1558M parameters).

\subsection{Input representation}
For our inputs, we add a start of sentence token (\emph{$[$CLS$]$} for BERT and \emph{<s>} for RoBERTa and GPT2). The two sentences of a given item are separated by a separator token, and the final masked word is denoted by \emph{$[$MASK$]$} for BERT and \emph{<mask>} for RoBERTa. GPT2 does not require a masked token. The special tokens are chosen based on the implementation of~\citet{wolf-etal-2020-transformers}.

\section{Results}



We begin by examining model prediction accuracy when attractors are semantically related to the critical background word or the target word. We define accuracy as percentage of instances in which models assign higher probability to the correct target than to any alternatives in the corresponding semantic set. This serves as the most direct test of the anticipated potential impact of our semantically related attractors, as it directly assesses whether presence of an irrelevant word (e.g., ``Indonesia''), which is semantically related to the critical background or target, and which invites a competing prediction (``Jakarta''), will cause models to prefer that competing prediction over the correct target.



Figure~\ref{fig:accbynum1} breaks down accuracy by number of attractors, for multiple- and single-entity settings. For the sake of space, we merge B-type and T-type attractors, which show largely similar patterns.\footnote{See Appendix Figures~\ref{fig:btypebynum-swapped}-\ref{fig:ttypebynum-swapped} for B-type and T-type results.} We see in Figure~\ref{fig:accbynum1} that addition of just a single semantically related attractor has clear impact on model performance, with models preferring the correct completion substantially less often than when no such attractors are present. \RL{} shows the strongest resistance to this effect, but still shows clear disruption from the first attractor.

As we insert additional attractors, for both conditions we see that rather than further hindering performance, for a couple of models accuracy actually improves. While we are not certain what drives this pattern, we speculate that a possible cause could be that models may learn to pay less attention to content that takes the form of lists. This is consistent with the fact that improvement with more attractors is mitigated in the multiple-entity setting, when attractors take the form of more complex statements, rather than lists of single words.  


\subsection{Impact on probabilities}

\begin{figure*}[ht]
    \includegraphics[width=.9\textwidth]{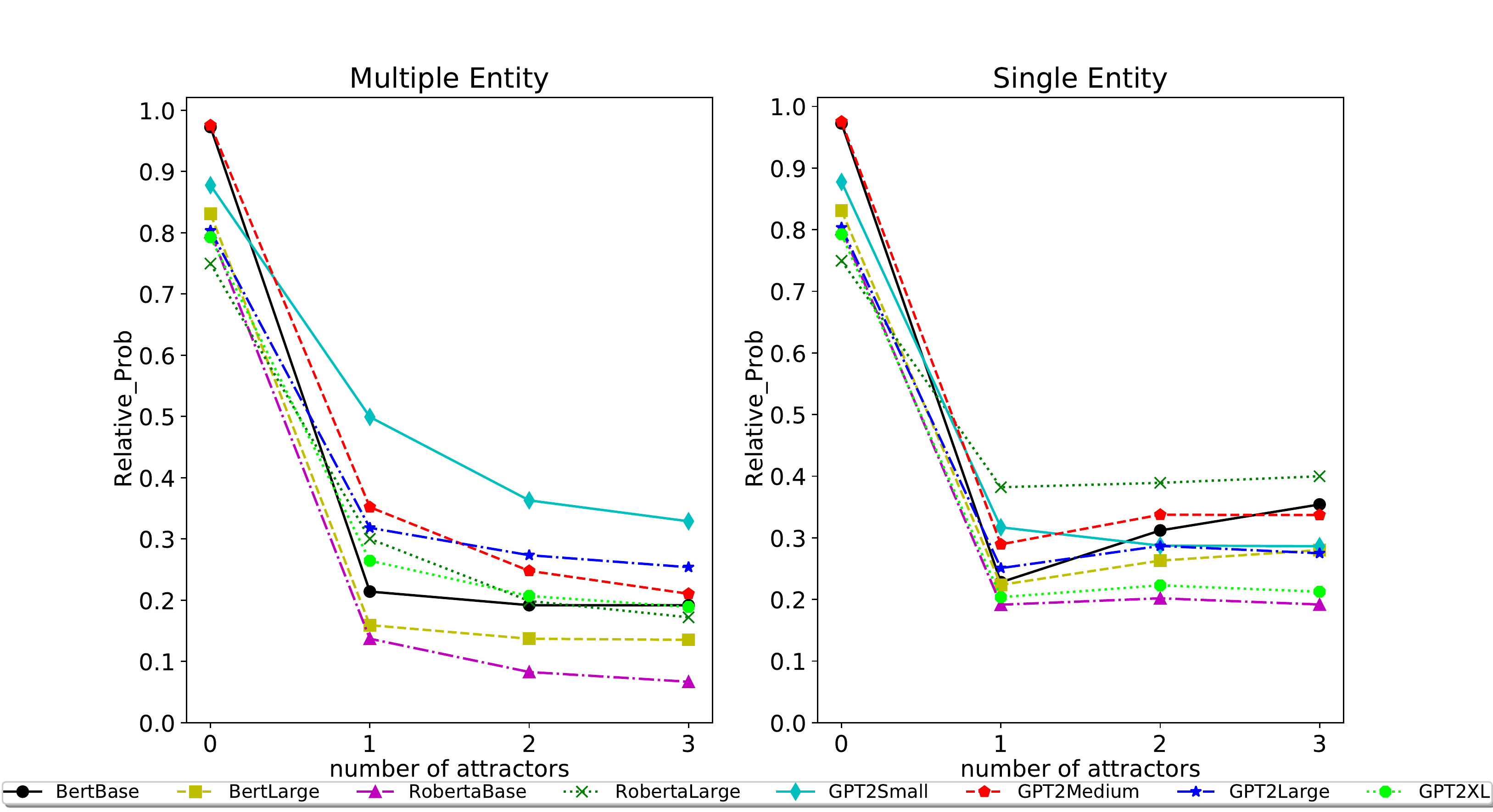}
    \centering
    \caption{Relative probability by number of attractors, with semantically related attractor type}
    \label{fig:probbynum1}
\end{figure*}

Since our definition of accuracy doesn't indicate impact of attractors on absolute target probabilities, we also examine how target probabilities change from base contexts to attractor contexts. We calculate relative probability as in Eq. \ref{relative_prob_eqn}, where $w$ is the candidate target word, $c_{attr}$ is the relevant attractor context, and $c_{base}$ is the base context:\\
\begin{equation}\label{relative_prob_eqn}\frac{P(w | c_{attr})}{P(w | c_{base})}\end{equation}
Figure~\ref{fig:probbynum1} shows these relative probabilities aggregated by number of attractors.\footnote{Values in the zero-attractor condition are not always 1 because the zero-attractor condition includes variants of the base context using the additional, semantically unrelated intervening material described in Section~\ref{sec:datcon}.} Consistent with the accuracy results, we see that in both settings, addition of just one semantically related attractor causes a dramatic drop in probability of the target relative to its base context level. This effect is especially uniform in the single-entity setting---in the multiple-entity setting, \GS{} shows less dramatic impacts with the first attractor. Also in keeping with the accuracy results, addition of further attractors does comparatively little damage beyond that of the first attractor, with relative probabilities remaining fairly stable with more attractors in the single-entity setting, and continuing to reduce, but very gradually, in the multiple-entity setting. 



\subsection{Semantically unrelated attractors}

The prior sections show that semantically related attractors, even if they are irrelevant for a prediction, have clear impacts on model outputs. To what extent is this effect driven by the relationship between the attractors and the critical sentence components? In this section we show results of adding attractors that are meaningful, but unrelated to the critical background word or the target. Examples are given in Table~\ref{tab:data}, and more in Appendix Table~\ref{tab:Unrelateddata}.

Our above definition of accuracy becomes less relevant with unrelated attractors, since accuracy was defined in terms of competition within semantic sets. Appendix Figure~\ref{fig:accbynum-neutralDistractor} shows accuracy results for semantically unrelated items, and confirms that models remain at very high accuracy when attractors are not semantically related to our chosen sets. Notably, however, there is a gradual but non-trivial drop in accuracy for most models, suggesting that in some cases even these semantically unrelated attractors are enough to confuse models into preferring an incorrect, semantically related completion.

\begin{figure*}[ht]
    \includegraphics[width=.9\textwidth]{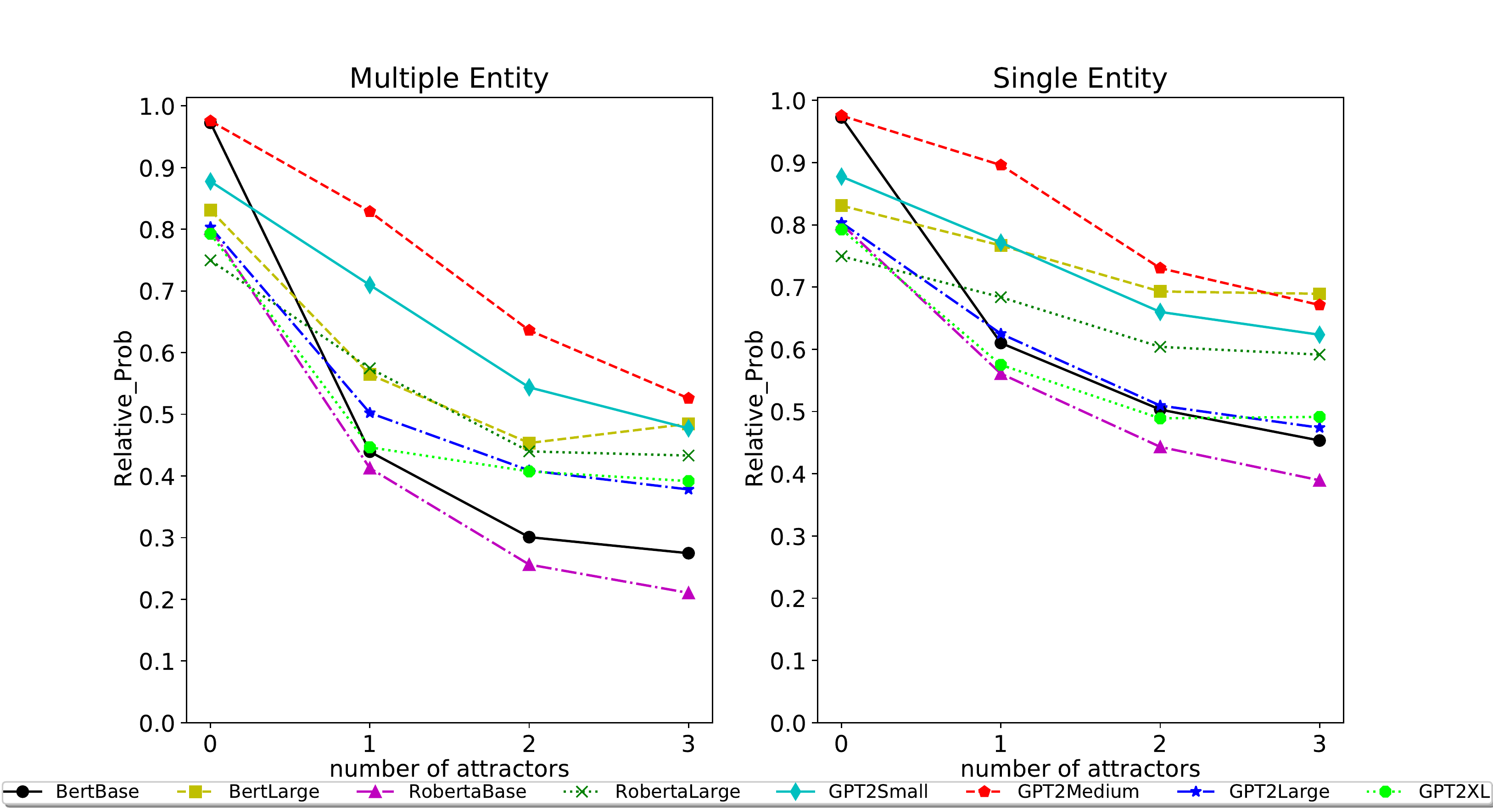}
    \centering
    \caption{Relative probability by number of attractors, with semantically unrelated attractor type}
    \label{fig:probbynum-neutral}
\end{figure*}



More appropriate to examine here are the impacts of the semantically unrelated attractors on models' target probabilities relative to the base probability. These are shown in Figure~\ref{fig:probbynum-neutral}. We see that unrelated attractors do have non-trivial impact on target probabilities, with fairly smooth reduction in relative probability as more attractors are added. This suggests that whether or not attractors are semantically related to the critical background fact or target, their presence still affects models' confidence about the correct target, despite the fact that these attractors are irrelevant to the prediction.

Importantly, a key difference that we see between semantically related attractors and semantically unrelated attractors is in the dramatic dip---in both accuracy and relative probability---with addition of the first semantically related attractor. This dip is missing when attractors are semantically unrelated to the critical background or target. It seems, then, that when the context contains an attractor occupying a similar location in the semantic space relative to facts being invoked for prediction, models are highly sensitive to even a single such item, and models' ability to make correct predictions is significantly hindered. Subsequent semantically related items show greatly diminished impacts, suggesting that addition of a single semantically related word achieves roughly ceiling impact on predictions. By contrast, if instead the distracting material occupies a more distant position in the semantic space, this material can still hinder models' predictions, but it does so less dramatically. The continually increasing impact with larger numbers of unrelated attractors also suggests that unlike the effects of semantically related attractors, the effects of semantically unrelated attractors on prediction are more gradual and additive in nature. 

Taken together, the results presented in these sections suggest that model predictions are significantly informed by superficial contextual cues, rather than by robust representations based on meaning of prior context. Differences between attractor types furthermore suggest that models rely heavily on coarse-grained semantic similarity cues to identify relevant context words for prediction.



\section{Varying position of information}

In this section we explore the effects of further varying entity and attractor position, to better understand how the models utilize positional cues.

\subsection{Separating key entity and critical fact}

In the previous section, all attractors occur after the critical background fact has been stated. What will happen if attractors fall between the key entity and the critical fact about that entity? Inserting attractors between these two components allows us to test the hypothesis that models might rely on proximity between the key entity and the critical fact in order to form a link (albeit, based on the above results, a brittle link) between the two. We use the same attractor sets, and simply adjust our templates to change the position of the attractors. We focus on semantically related attractors for this section. Appendix Table~\ref{tab:swappeddata} gives some examples.

\begin{figure*}[ht]
    \includegraphics[width=.9\textwidth]{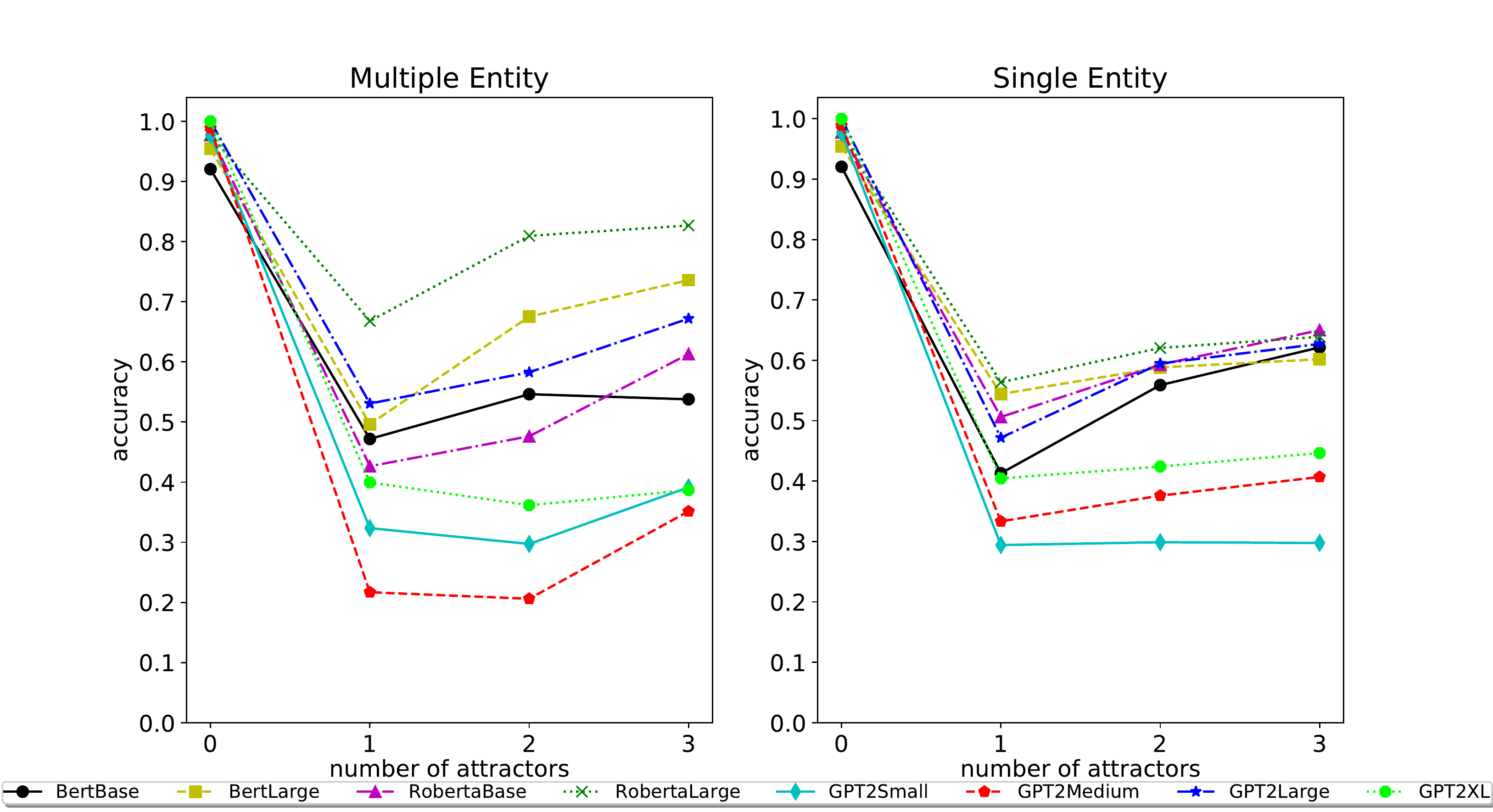}
    \centering
    \caption{Accuracy by number of attractors, with attractors intervening between key entity and critical fact}
    \label{fig:accbynum-swapped}
\end{figure*}
The results of this analysis are shown in Figure~\ref{fig:accbynum-swapped}. On the whole, the patterns are quite similar to when attractors occur after the critical background fact. In both single- and multiple-entity settings we see the clear dip in accuracy after a single attractor. The multiple-entity setting produces a bit more spread between models, suggesting in particular that \RL{} thrives---in fact, improves---despite the key entity being separated from the critical fact, while \GM{} performs very poorly. We also see that many models improve with more attractors, suggesting again some effect in which models may learn to down-weight content in lists.

On the whole, the results suggest that regardless of whether attractors intervene between the key entity and key fact, or between the key fact and the target position, outcomes are similar: just a single semantically related attractor in the context will significantly disrupt models' ability to make a correct prediction. These results also suggest that models don't put heavy reliance on proximity between the key entity and the critical fact. While this could suggest that the models are robust in forming entity-fact links, it could also indicate that the models aren't really forming those links at all.

\subsection{Varying key entity position}

In all of our test items up to this point, the key entity has also been the first entity mentioned. In this section we test the impact of prompting a prediction about an entity that is not the first mention. We do this by taking our existing multiple-entity items, and adapting them so that the entity queried at the target position is one of the later-mentioned entities, rather than the first-mentioned entity (e.g. \emph{Sebastian lives in France, and Rowan lives in Indonesia. The capital of Rowan's country is $[$MASK$]$}).


\begin{figure*}[ht]
    \includegraphics[width=.9\textwidth]{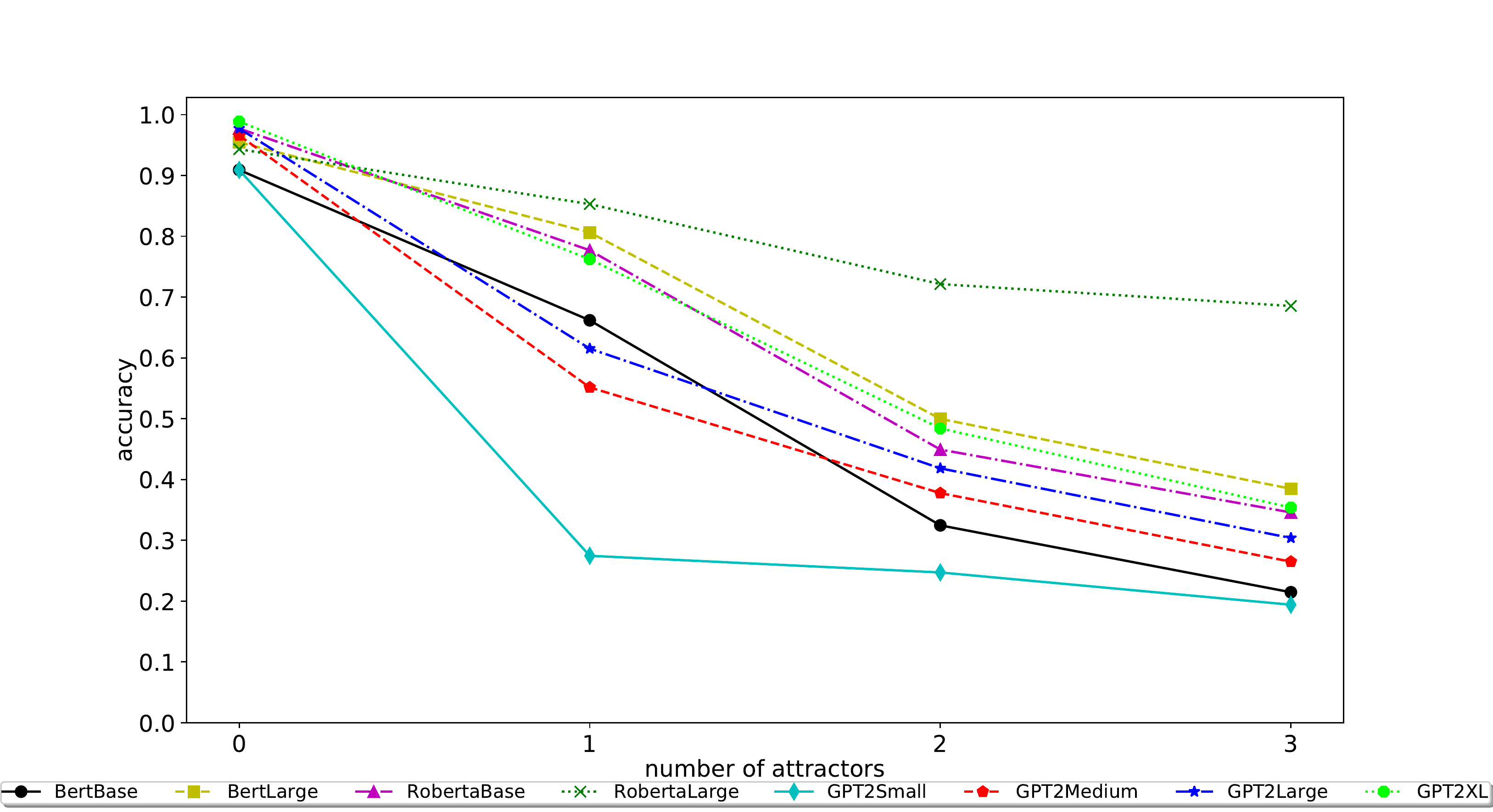}
    \centering
    \caption{Accuracy by number of attractors, with key entity occurring later in sentence}
    \label{fig:accbynum-entvar}
\end{figure*}

Figure~\ref{fig:accbynum-entvar} shows the results. We see that, with the exception of \GS{}, the dramatic dip on the first attractor is no longer present. Instead, we see steady decrease in accuracy with more attractors. What causes this change? The major difference in the one-attractor condition here, relative to our previous settings, is that the key entity and critical background fact now both occur closer to the target position than the attractor does. Thus, the observation that this more distant attractor has less effect on predictions suggests that in addition to semantic similarity, model predictions are also strongly impacted by recency---that is, a semantically related attractor has dramatic impact on prediction accuracies, but only if it occurs after the critical background entity. This pattern of results also suggests that models may learn to link entity mentions primarily with words occurring after those mentions, and not before. It is easy to see how such a heuristic may arise, as entity descriptors will more frequently follow the entity mention---however, this assumption is not foolproof, and it does not help with distinguishing relevant versus irrelevant words that occur after the entity mention. In aggregate, these results further support the conclusion that superficial cues exert significant influences on model predictions, with these latter results suggesting a key role for word position relative to the key entity.

We note that \RL{} again stands out as by far the most robust to these attractor effects, though it shows a notable decrease nonetheless.



\section{Discussion}

The experiments above were designed to do two things. The first purpose was to test whether pre-trained LMs show evidence of robustly processing and storing new facts from input context. To test this, we inserted irrelevant, distracting content in addition to critical context information, to see whether model predictions would be affected. The results of these experiments show clearly that model predictions are indeed impacted by this irrelevant content, rather than remaining consistent on the basis of the critical contextual information. 

The second goal of these experiments was to explore the specific types of cues affecting model predictions, by varying the nature of distracting content. Through these manipulations, we find influences of both semantic similarity and word position as superficial cues informing model predictions. In particular, we find that insertion of a single semantically related attractor has a dramatic effect on predictions, with the first such attractor reaching roughly ceiling impact. Unrelated attractors also influence predictions, with more gradual, additive impacts. Effects of semantic similarity interact additionally with effects of relative word position, with semantically related attractors showing much reduced influence when occurring prior to the key entity mention. These trends suggest that models rely on heuristics involving semantic similarity to critical words, and relative position of entity and descriptor words, for determining which elements of context are relevant for a prediction.

While it should not come as a shock that language models use superficial contextual cues for prediction, findings of this kind must serve as reality checks as we consider the capacity of these models for language ``understanding''. We have made the simple assumption that ``understanding'' will involve robustly representing and retaining information from prior context, and generating predictions accordingly. The results presented here suggest that this criterion is not met. Additionally, the results of these experiments take steps toward understanding the precise strategies that models do use in generating predictions, beginning to sketch out a picture in which models rely substantially on coarse-grained semantic similarity and word position cues to identify relevant contextual words.

We do note that of the models tested, \RL{} frequently distinguishes itself as least susceptible to our attractors---though it shows disruption all the same. Since this model is pre-trained on a larger dataset than other models tested here, size of pre-training data is a likely contributor to the model's superior performance. Additionally, we find that within a given training regime, larger models mostly perform more robustly than smaller models, suggesting that model size is also a contributor in interaction with factors like training data size. We leave for future work the investigation of whether the comparatively strong performance of \RL{} reflects truly more robust representations in that model---or use of superficial cues not yet targeted in the present work.

These findings also have implications for studying mechanistic connections between pre-trained language models and language processing in humans. Studies of human sentence processing have shown comparable susceptibility to interference from irrelevant context elements, depending on semantic and syntactic properties~\cite{van2007interference,parker2017reflexive,dillon2013contrasting}. Systematic comparison of interference effects in humans and in language models stands to shed light on mechanistic similarities and differences in the ways that these language processing systems handle information from prior context.

\section{Conclusion}

We have presented results manipulating inputs of pre-trained LMs, to test the ability of such models to represent and retain information conveyed by input text. Our results show that though models may appear to handle information correctly in simple settings, these correct predictions are easily broken by insertion of distracting material in the context. Systematic manipulation of the distracting content further indicates key roles for semantic similarity and relative word position in models' selection of relevant contextual cues for prediction. Overall, the results suggest that LM predictions are driven more by coarse-grained superficial cues than by extraction of robust meaning information from context. The results serve as a reality check for considerations of the extent to which LMs ``understand'' their input, and lay groundwork to understand the mechanisms that do drive predictions in these models.

\section*{Acknowledgments}

We would like to thank three anonymous reviewers for valuable feedback on this paper. We also thank members of the University of Chicago CompLing Lab for helpful discussion. This material is based upon work supported by the National Science Foundation under Award No.~1941160.


\bibliography{anthology,custom}
\bibliographystyle{acl_natbib}

\appendix
\section{Appendix}\label{sec:appendix}
\begin{table*}[ht]
\centering
\begin{tabular}{p{.8\textwidth}|p{.2\textwidth}}
\toprule

\textbf{Base context} & target \\
\midrule
\textbf{Sebastian} lives in \textbf{France}. The capital of Sebastian's country is \_\_\_\_ & Paris\\
 \bottomrule
 \textbf{Rowan} lives in \textbf{Chile}. The capital of Rowan's country is \_\_\_\_ & Santiago\\
 \bottomrule
  \textbf{Rowan} lives in \textbf{China}. The capital of Rowan's country is \_\_\_\_ & Beijing\\
 \bottomrule
 \textbf{Rowan} lives in \textbf{Finland}. The capital of Rowan's country is \_\_\_\_ & Helsinki\\
 \bottomrule
 \textbf{Rowan} lives in \textbf{Indonesia}. The capital of Rowan's country is \_\_\_\_ & Jakarta\\
 \bottomrule
  \textbf{Jake} lives in \textbf{Poland}. The capital of Jake's country is \_\_\_\_ & Warsaw\\
 \bottomrule
 \textbf{Jake} works as a \textbf{florist}. For his job, Jake sells \_\_\_\_ & flowers\\
 \bottomrule
  \textbf{Jake} works as an \textbf{optician}. For his job, Jake sells \_\_\_\_ & glasses\\
 \bottomrule
 \bottomrule
   \textbf{Jake} works as a \textbf{baker}. For his job, Jake sells \_\_\_\_ & bread\\
 \bottomrule
 \textbf{Daniel} works as a \textbf{butcher}. For his job, Daniel sells \_\_\_\_ & meat\\
 \bottomrule
 \textbf{Daniel} works as a \textbf{fisherman}. For his job, Daniel sells \_\_\_\_ & fish\\
 \bottomrule
 \textbf{Daniel} works as a \textbf{painter}. For his job, Daniel sells \_\_\_\_ & paintings\\
 \bottomrule
 \textbf{Daniel} visited the \textbf{Taj Mahal}. The country Daniel traveled to was \_\_\_\_ & India\\
 \bottomrule
 \textbf{Daniel} visited the \textbf{Pyramid of Giza}. The country Daniel traveled to was \_\_\_\_ & Egypt\\
 \bottomrule
 \textbf{Jack} visited the \textbf{Eiffel Tower}. The country Jack traveled to was \_\_\_\_ & France\\
 \bottomrule
 \textbf{Jack} visited the \textbf{Tower of Pisa}. The country Jack traveled to was \_\_\_\_ & Italy\\
 \bottomrule
  \textbf{Jack} visited the \textbf{Machu Picchu}. The country Jack traveled to was \_\_\_\_ & Peru\\
 \bottomrule
 \textbf{Jack} visited the \textbf{Kremlin}. The country Jack traveled to was \_\_\_\_ & Russia\\
 \bottomrule
  \bottomrule
 \textbf{Jack} played \textbf{football}. In his game, Jack scored a \_\_\_\_ & touchdown\\
 \bottomrule
 \textbf{Jack} played \textbf{baseball}. In his game, Jack scored a \_\_\_\_ & run\\
 \bottomrule
 \textbf{Daniel} played \textbf{soccer}. In his game, Daniel scored a \_\_\_\_ & goal\\
 \bottomrule
 \textbf{Sebastian} played \textbf{cricket}. In his game, Sebastian scored a \_\_\_\_ & century\\
 \bottomrule
\end{tabular}
\caption{Base context for the dataset}\label{tab:basedata}
\end{table*}
\begin{table*}[ht]
\centering
\begin{tabular}{p{.8\textwidth}|p{.2\textwidth}}
\toprule

\textbf{Context} & target \\
\midrule
John lives in Chile and \textbf{writes poetry}. The capital of John's country is \_\_\_\_ & Santiago\\
 \bottomrule
 John lives in Chile, \textbf{writes poetry, and drives a car}. The capital of John's country is \_\_\_\_ & Santiago\\
 \bottomrule
  John lives in Chile, \textbf{writes poetry, and drives a car}.	The capital of John's country is \_\_\_\_ & Santiago\\
 \bottomrule
 John lives in Chile, \textbf{writes poetry, drives a car, and slept late last week}.	The capital of John's country is \_\_\_\_ & Santiago\\
 \bottomrule
 John works as a florist and \textbf{Jack writes poetry}.	For his job, John sells \_\_\_\_ & flowers\\
 \bottomrule
 Jake visited the Eiffel Tower, \textbf{Rowan drives a car, and Jack sits by the lake}. The country Jake traveled to was \_\_\_\_ & France\\
  \bottomrule
 Sebastian played football, \textbf{writes poetry, slept late last week, and sits by the lake}. In his game, Sebastian scored a \_\_\_\_ & touchdown \\
 \bottomrule
 \end{tabular}
\caption{Examples from semantically unrelated attractors}\label{tab:Unrelateddata}
\end{table*}

\begin{table*}[ht]
\centering
\begin{tabular}{p{.8\textwidth}|p{.2\textwidth}}
\toprule
\textbf{Context} & target \\
\midrule
Daniel knows that Jack lives in Beijing and he himself lives in Chile.	The capital of Daniel's country is \_\_\_\_ & Santiago\\
 \bottomrule
Daniel knows that Jake likes to buy glasses and Rowan likes to buy meat and he himself works as a florist.	For his job, Daniel sells \_\_\_\_ & flowers\\
 \bottomrule
Joe wants to visit the Eiffel Tower, the Pyramid of Giza, and the Machu Picchu and has only visited the Taj Mahal.	The country Joe traveled to was \_\_\_\_ & India\\
 \bottomrule
 Rowan knows that his friends scored a goal and a century and he himself played football. In his game, Rowan scored a \_\_\_\_ & touchdown\\
  \bottomrule
 \end{tabular}
\caption{Examples from separating key entity and critical fact}\label{tab:swappeddata}
\end{table*}
\begin{figure*}[ht]
    \includegraphics[width=.9\textwidth]{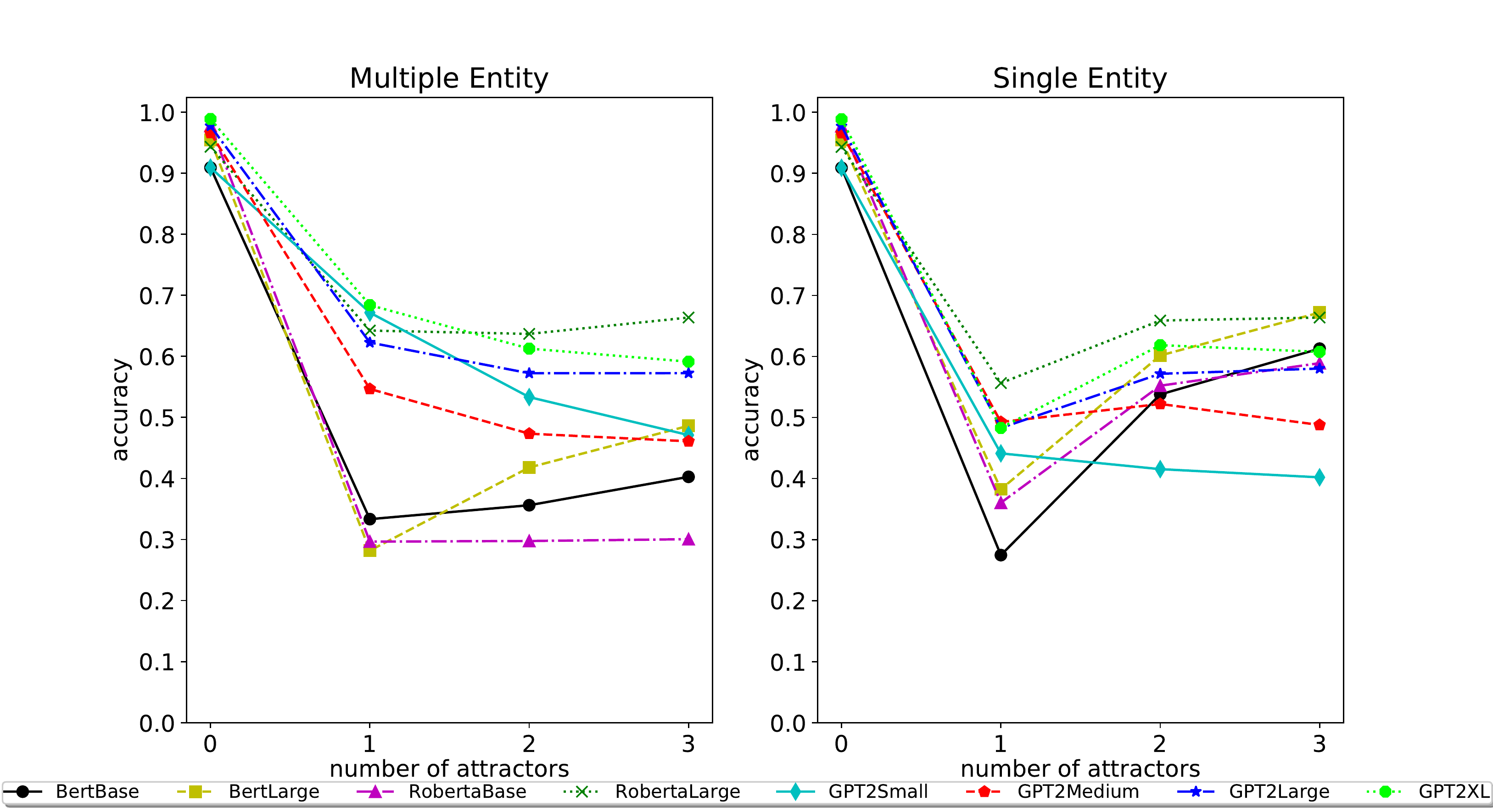}
    \centering
    \caption{Accuracy vs number of attractors, with B type}
    \label{fig:btypebynum-swapped}
\end{figure*}

\begin{figure*}[ht]
    \includegraphics[width=.9\textwidth]{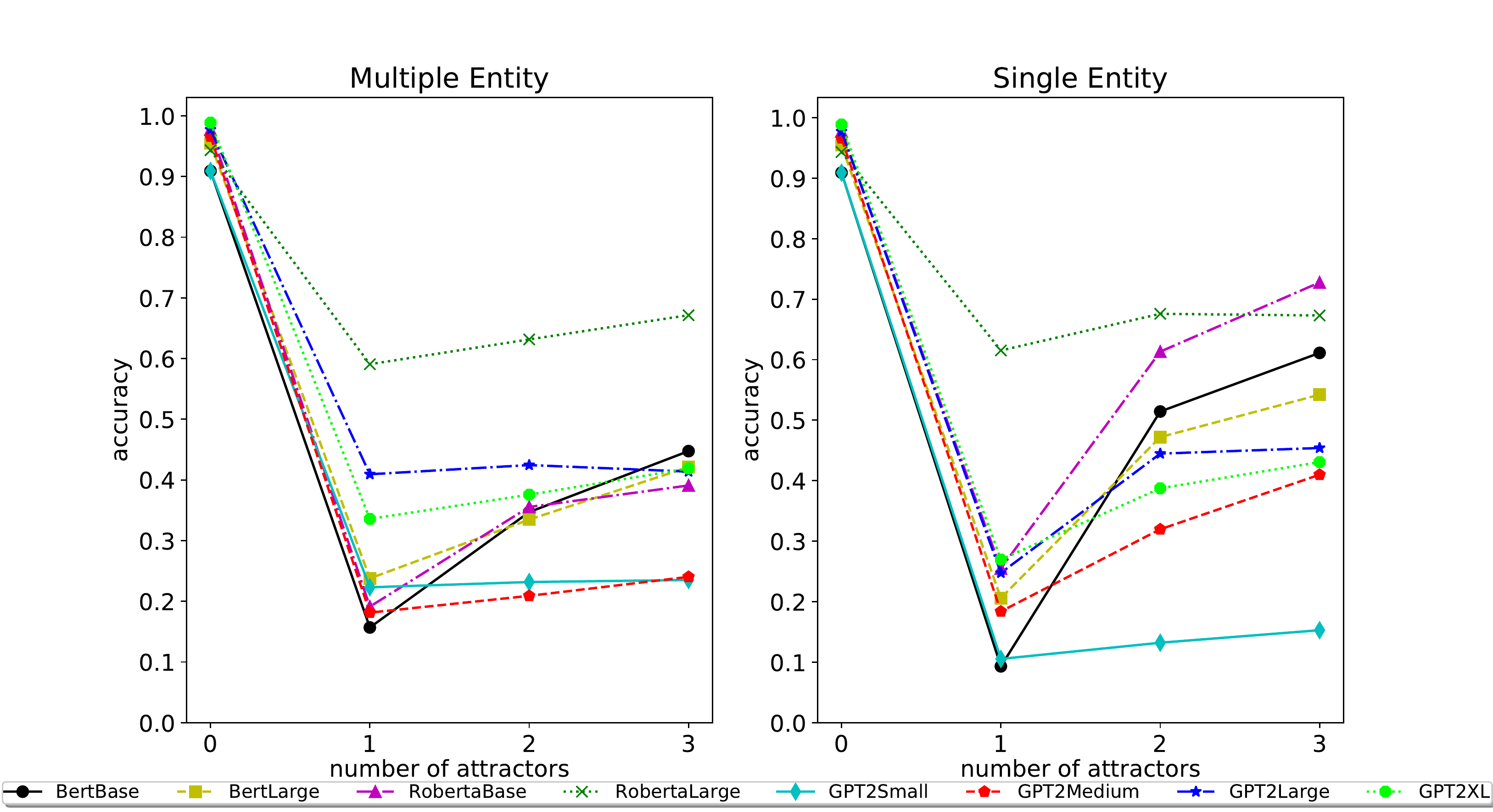}
    \centering
    \caption{Accuracy vs number of attractors, with T type}
    \label{fig:ttypebynum-swapped}
\end{figure*}

\begin{figure*}[ht]
    \includegraphics[width=.9\textwidth]{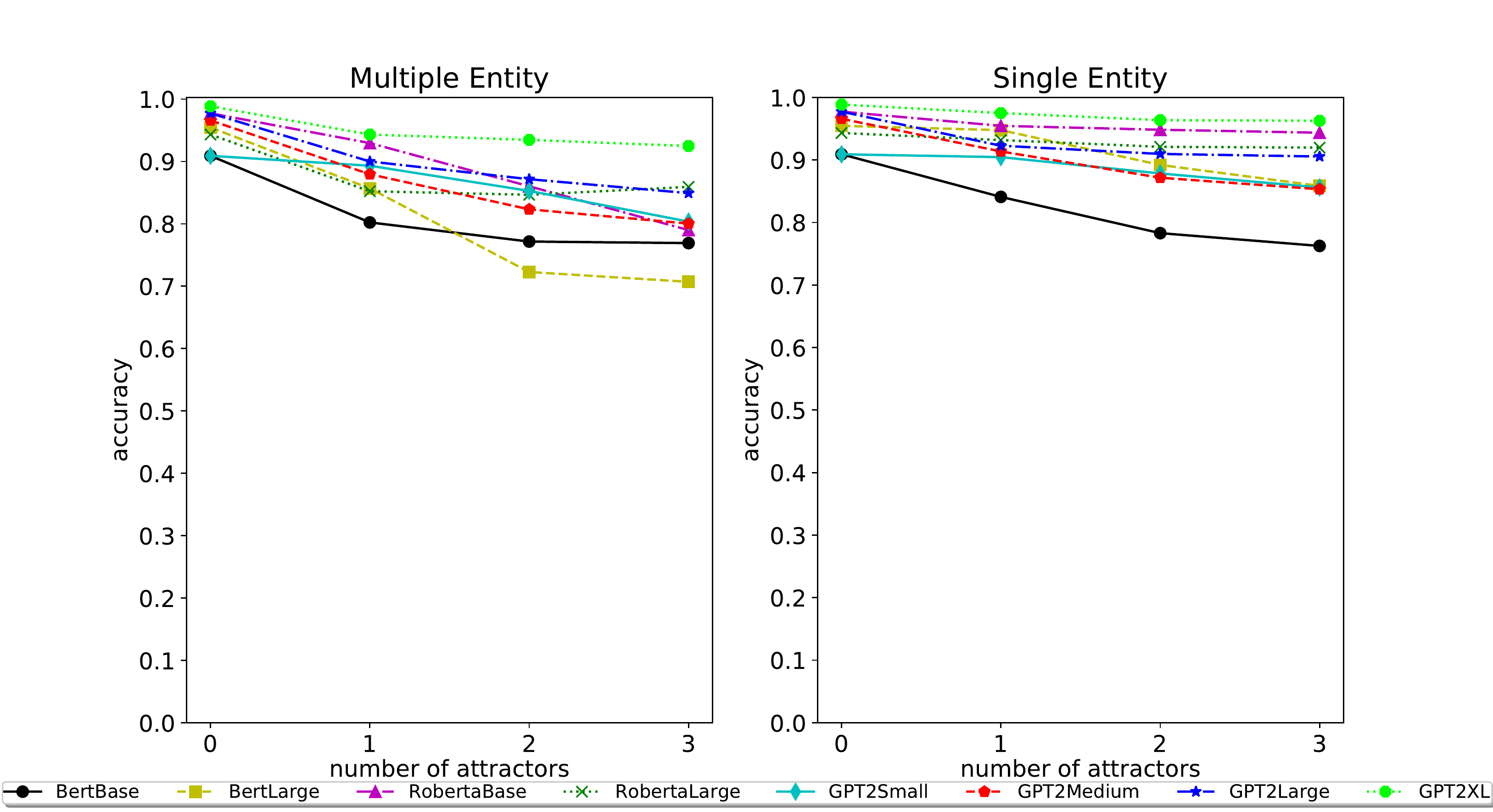}
    \centering
    \caption{Accuracy vs number of attractors with semantically unrelated attractors}
    \label{fig:accbynum-neutralDistractor}
\end{figure*}


\end{document}